\documentclass{article}

\usepackage{arxiv}

\usepackage[utf8]{inputenc} 
\usepackage[T1]{fontenc}    
\usepackage{hyperref}       
\usepackage{url}            
\usepackage{booktabs}       
\usepackage{amsfonts}       
\usepackage{nicefrac}       
\usepackage{microtype}      
\usepackage{lipsum}		
\usepackage{graphicx}
\usepackage{natbib}
\usepackage{doi}
\usepackage{multirow} 
\usepackage{amsmath}
\usepackage{graphicx}
\usepackage{array}
\title{Advancing Interpretability in Text Classification through Prototype Learning}

\date{} 					

\author{ \href{https://orcid.org/0009-0005-3204-0255}{\includegraphics[scale=0.06]{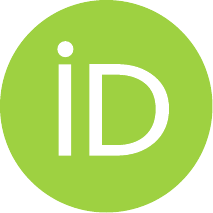}\hspace{1mm}Bowen Wei} \\
	Department of Computer Science\\
	George Mason University\\
	Fairfax, VA 22030 \\
	\texttt{bwei2@gmu.edu} \\
	\And
	 \href{https://orcid.org/0000-0002-3990-4774}{\includegraphics[scale=0.06]{orcid.pdf}\hspace{1mm}Ziwei Zhu} \\
	Department of Computer Science\\
	George Mason University\\
	Fairfax, VA 22030 \\
	\texttt{zzhu20@gmu.edu} \\
}

\begin{document}
\maketitle

\begin{abstract}
Deep neural networks have achieved remarkable performance in various text-based tasks but often lack interpretability, making them less suitable for applications where transparency is critical. To address this, we propose ProtoLens, a novel prototype-based model that provides fine-grained, sub-sentence level interpretability for text classification. ProtoLens uses a Prototype-aware Span Extraction module to identify relevant text spans associated with learned prototypes and a Prototype Alignment mechanism to ensure prototypes are semantically meaningful throughout training. By aligning the prototype embeddings with human-understandable examples, ProtoLens provides interpretable predictions while maintaining competitive accuracy. Extensive experiments demonstrate that ProtoLens outperforms both prototype-based and non-interpretable baselines on multiple text classification benchmarks. Code and data are available at \url{https://anonymous.4open.science/r/ProtoLens-CE0B/}.
\end{abstract}


\section{Introduction}
\label{sec:intro}
Deep neural networks (DNNs) have achieved remarkable success in various natural language processing tasks, including text classification~\citep{kowsari2019text}, sentiment analysis~\citep{medhat2014sentiment}, and question answering~\citep{allam2012question}. However, their black-box nature presents significant challenges for interpretability, limiting their use in high-stakes applications where transparency, user trust, and accountability are paramount~\citep{castelvecchi2016can, rudin2019stop}. While post-hoc explanation methods attempt to address this~\citep{jacovi2018understanding, mishra2017local}, they often lack faithfulness and consistency in explaining predictions~\citep{rudin2019stop}. In contrast, inherently interpretable models guarantee transparency, facilitating understanding and trust in model outputs~\citep{molnar2020interpretable}.

Among various approaches aimed at enhancing model interpretability, prototype-based methods have emerged as a prominent line of research. It enables models to generate predictions by comparing inputs to prototypical examples, similar to human reasoning that relies on analogies to familiar cases \cite{dong2018few,hong2023protorynet, zhang2023learning, ming2019interpretable, gautam2022protovae, ming2019protosteer, rajagopal2021selfexplain, sourati2023robust}. Such prototype-based models provide an intuitive form of interpretability, facilitating an understanding of model predictions through direct reference to interpretable examples. For instance, in a movie review classification task, a prototype might represent a review like "This movie was amazing, with stunning visuals and a gripping storyline", which the model would use to classify new reviews with similar sentiments. The model can explain its classification of a new review by showing how closely it matches this prototypical example. 


\begin{figure}[t]
    \centering
    \includegraphics[width=0.8\linewidth]{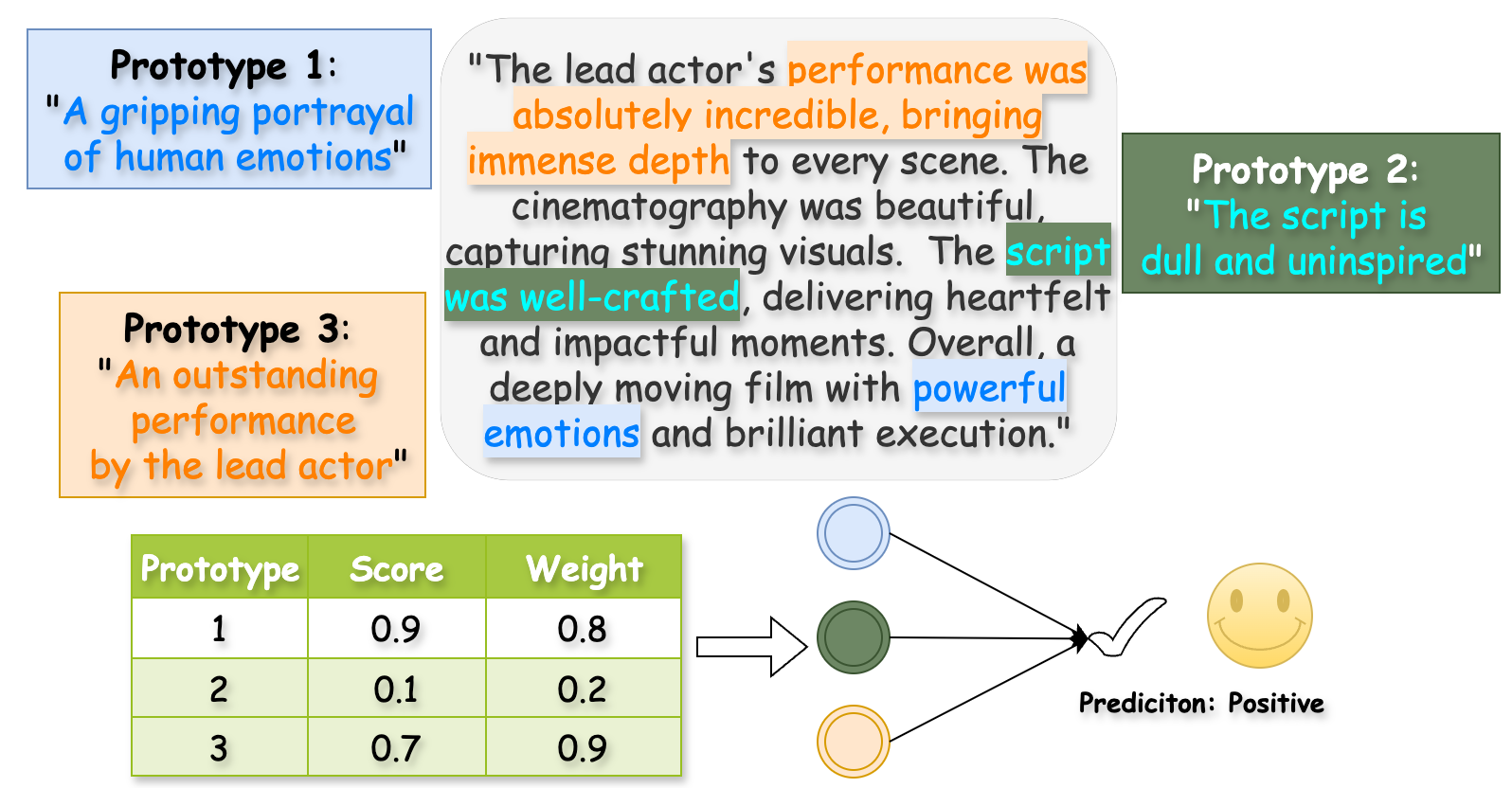}
    \caption{ Interpretable classification by ProtoLens. }
    \label{fig:Intro}
\end{figure}

Despite the potential of prototype-based models for enhancing interpretability, existing approaches encounter significant limitations in text-based applications~\citep{hong2023protorynet, ming2019interpretable}. Typically, these models define prototypes at the instance/sentence level, which often lacks the granularity needed for fine interpretability in complex or lengthy texts. For example, in a movie review like "The visuals were stunning, but the plot was too predictable", a sentence-level prototype might only capture the general sentiment of the review, missing the nuance that the visuals were positive, while the plot had negative aspects. This coarse granularity makes it challenging to provide insightful explanations when different sentiments or nuances coexist within a single text. In contrast, more fine-grained prototype modeling, such as sub-sentence level, is crucial for delivering detailed interpretative insights, allowing the model to explain specific aspects of the text, like "stunning visuals" or "predictable plot".

To address this challenge,  a novel prototype-based model ProtoLens is designed for finer-grained interpretability. ProtoLens builds on the concept of prototypical learning but extends it in key ways that make it better suited for handling the complexities inherent in textual data. The general reasoning process of ProtoLens is illustrated by the example in Figure~\ref{fig:Intro}: ProtoLens leverages three prototypes related to "emotion", "performance", and "script", and extracts prototype-specific text units from the input review. Based on extracted units, Prototype 1 and 2 are activated and thus positive prediction is derived.

There are two core modules in the proposed ProtoLens. First, for a specific prototype, the \textbf{Prototype-aware Span Extraction} module employs a Dirichlet Process Gaussian Mixture Model (DPGMM)~\citep{gorur2010dirichlet, rasmussen1999infinite} to extract relevant text spans in inputs for the purpose of model prediction and interpretation. This module enables sub-sentence extraction and offers a more accurate and finer-grained extraction of text spans for certain prototypes. Second, we devise the \textbf{ Prototype Alignment} mechanism, which adaptively aligns the learned prototype embeddings with representative data samples throughout training. By this, we ensure that learned prototypes are semantically reasonable and effective for interpretation.




Extensive experiments demonstrate that ProtoLens not only outperforms competitive baselines on multiple text classification benchmarks but also provides more intuitive and user-friendly explanations for its predictions. 

\begin{figure*}[t]
    \centering
    \includegraphics[width=0.8\linewidth]{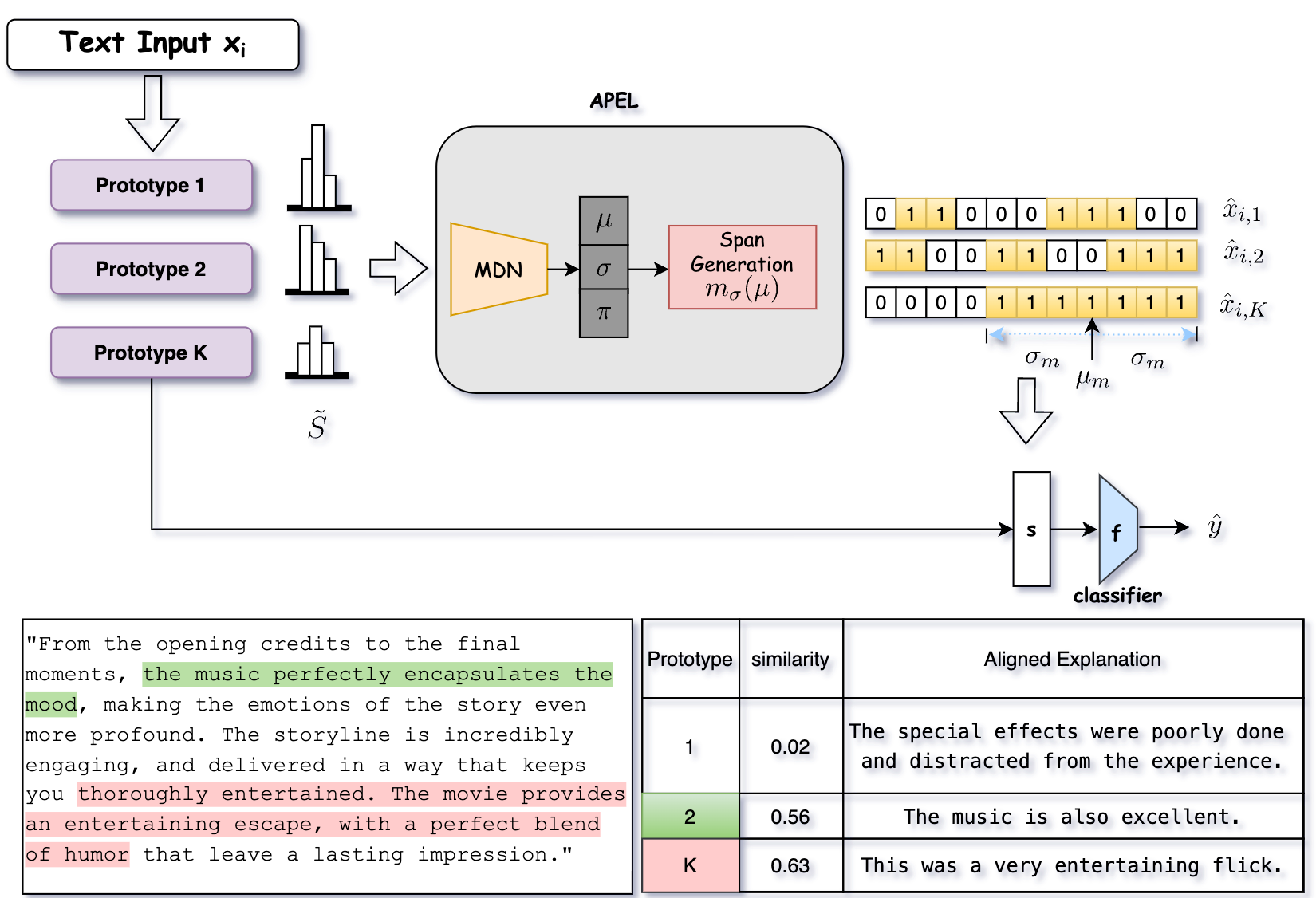}
    \caption{Model Structure. }
    \label{fig:model}
\end{figure*}

\begin{figure*}[t]
    \centering
    \includegraphics[width=0.6\linewidth]{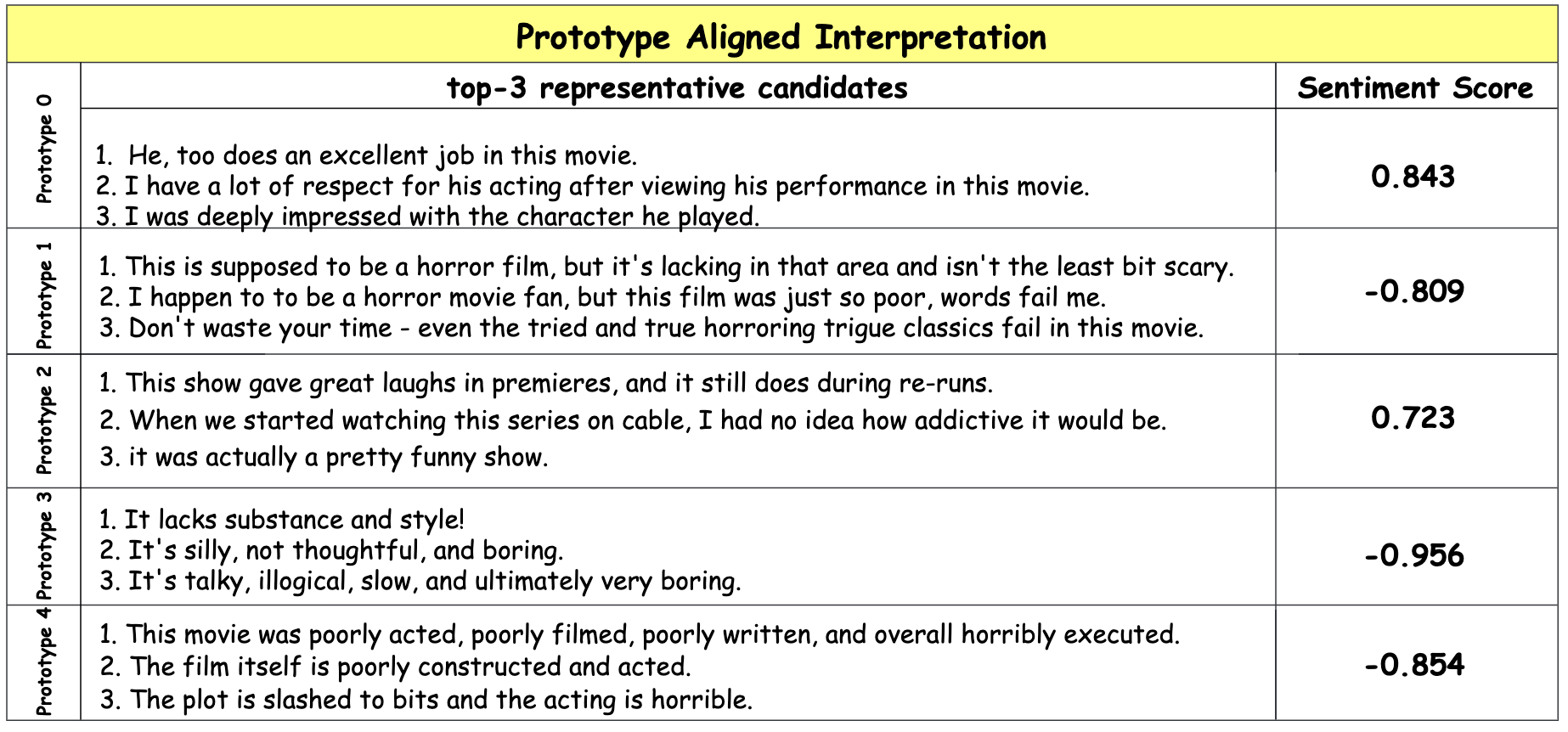}
    \caption{Sampled Aligned interpretation of prototypes with corresponding text sentences.}
    \label{fig:aligned_interpretation}
\end{figure*}

\begin{figure*}[t]
    \centering
    \includegraphics[width=0.8\linewidth]{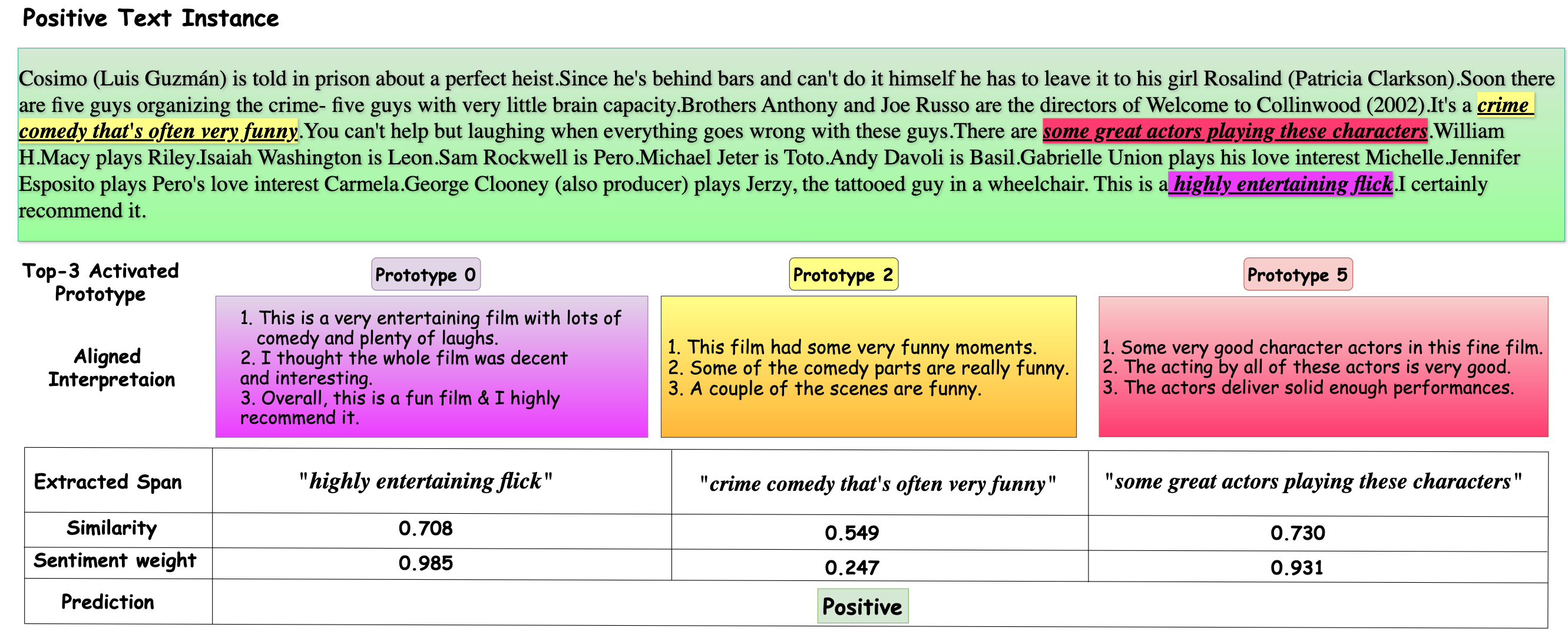}
    \caption{Case study of a positive class text instance.}
    \label{fig:case_pos}
\end{figure*}

\begin{figure*}[t]
    \centering
    \includegraphics[width=0.8\linewidth]{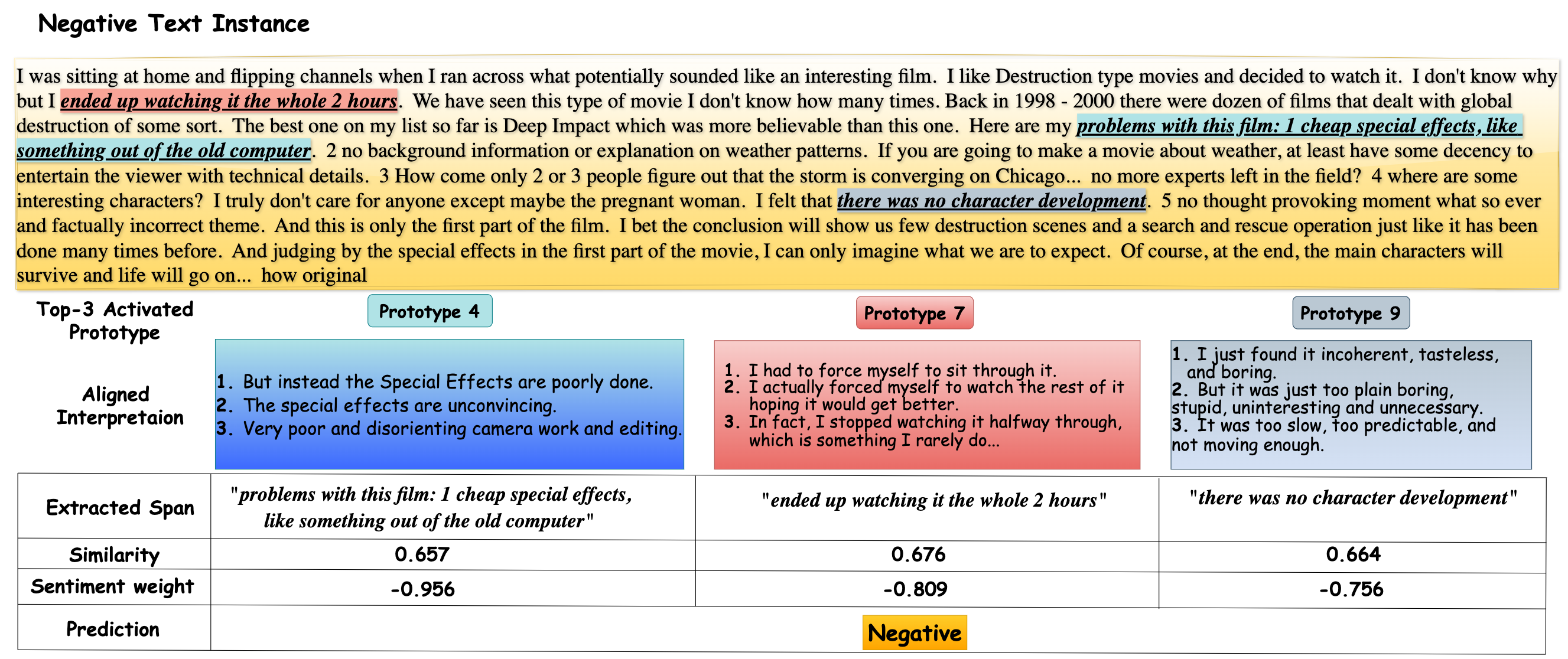}
    \caption{Case study of a negative class text instance.}
    \label{fig:case_neg}
\end{figure*}

\section{Related Work}

\paragraph{Post-hoc Explanations.} 

Several post-hoc methods interpret DNN models by analyzing gradients or neuron activations, such as Integrated Gradients \citep{sayres2019using, qi2019visualizing}, DeepLift \citep{li2021deep}, and NeuroX \citep{nalls2015neurox}. \citet{tsang2018can} proposed a hierarchical method to capture interaction effects, later adapted by \citet{jin2019towards} for text classification. In sentiment analysis, contextual decomposition \citep{murdoch2018beyond} identifies sentiment words and their contributions. Attention-based models, such as \citet{bahdanau2014neural, rocktaschel2015reasoning}, analyze attention weights, though \citet{jain2019attention} question their explanatory power. 

\paragraph{Prototype-based Deep Neural Networks.} Prototype-based deep neural networks enhance interpretability by using prototypes as intuitive references for decision-making, a concept with roots in traditional models \citep{sorgaard1991evaluating, fikes1985role, kim2014bayesian}. ProtoPNet \citep{chen2019looks} pioneered the integration of prototypical reasoning into deep learning for image classification, while ProtoVAE \citep{gautam2022protovae} advanced this with a variational autoencoder for diverse, interpretable prototypes. In text classification, ProSeNet \citep{ming2019interpretable} introduced prototype-based reasoning, later extended by ProtoAttend \citep{arik2020protoattend} with attention mechanisms for dynamic prototype selection. ProtoryNet \citep{hong2023protorynet} further evolved the field by modeling prototype trajectories for improved interpretability across multiple domains.

Unlike previous methods, our approach embeds interpretability directly into the model, offering sub-sentence level interpretability, providing more granular insights than word- or sentence-level approaches.

\section{Method}
To deliver inherently interpretable predictions at a fine-grained level, we introduce \textbf{ProtoLens}, a prototype-based interpretable neural network. ProtoLens is designed to overcome two primary challenges: (C1) How to effectively extract text spans associated with a given prototype to provide interpretable predictions? and (C2) How to ensure learned prototypes are semantically reasonable and effective for interpretation? To address C1, we propose a Prototy-Aware Span Extraction module, which extracts most relevant text spans for prototypes by a Dirichlet Process Gaussian Mixture Model. To address C2, we design a Prototype Alignment mechanism to adaptively align prototype embeddings to representative data samples through training. The overall model architecture is illustrated in Figure~\ref{fig:model}.



\noindent \textbf{Model Structure.} Given a corpus of textual data $\mathcal{D} = \{(x_i, y_i)\}$, where $i = 1, \ldots, N$, each instance $x_i$ is associated with a label $y_i \in \mathcal{Y}$, our model processes the text through a text encoder, such as BERT, $r: \mathcal{X} \to \mathbb{R}^d$, where $\mathcal{X}$ represents the space of text inputs and $d$ is determined by the encoder architecture.

The text instance is passed through an \textbf{Prototype-aware Span Extraction} module, containing a set of trainable prototypes $\mathcal{P} = \{\mathbf{p}_k \in \mathbb{R}^d : k = 1, \dots, K\}$, where each prototype is represented by a learnable embedding, and $K$ is the number of prototypes specified by the user. The model will deliver classifications by comparing the input to these prototypes. For each prototype, we aim to identify relevant text spans in the input. Specifically, for prototype $k$ and input instance $x_i$, a span $\mathbf{R}_i^k = [R_{i, t}^k]_{t=1}^T$ is generated, where $R_{i, t}^k \in \{0, 1\}$ indicates whether the token at position $t$ is selected for prototype $k$. Each prototype $k$ has an independent span $\mathbf{R}_i^k$ for every instance $x_i$, capturing the most relevant regions pertinent to that specific prototype. The generated spans are then used to refine the token embeddings of $x_i$:

\begin{equation}
    \mathbf{Z}_i^k =  r(\mathbf{R}_i^k \odot x_i),
\end{equation}
where $\odot$ denotes element-wise multiplication. The refined embedding $\mathbf{Z}_i^k$ thus represents the most relevant portions of $x_i$ with respect to prototype $k$. 


The similarity between the refined representation $\mathbf{Z}_i^k$ and prototype $\mathbf{p}_k$ is then computed as $s_i^k = \text{RMSNorm}(cos(\mathbf{Z}_i^k, \mathbf{p}_k))$. The final prediction is computed via an interpretable model $f$ applied to the similarity vector $\mathbf{s}_i = [s_i^1, s_i^2, \dots, s_i^K]$: $\hat{y}_i = f(\mathbf{s}_i)$, where $\mathbf{s}_i$ captures the proximity to each prototype, quantifying their influence on the final prediction; and $f$ can be any interpretable models, such as decision tree or logistic regression. In this paper, we adopt logistic regression as $f$.

\subsection{Prototype-aware Span Extraction}
Given a text instance $x_i$, we divide it into parts $x_i = (c_t)_{t=1}^T$ using a sliding window approach, where $c_t$ denotes the $t$-th part and $T$ is the total number of parts. Each part is an n-gram span, with $n$ as a hyperparameter controlling the granularity of text partation. The text encoder processes each part $c_t \in x_i$ to produce an embedding $\mathbf{e}_t = r(c_t) \in \mathbb{R}^d$. The proximity $s_{t,k}$ between the part embedding $\mathbf{e}_t$ and the prototype embedding $\mathbf{p}_k$ is then measured using cosine similarity: $s_{t,k} = cos(\mathbf{e}_t, \mathbf{p}_k)$.

The intermediate output of the module is the similarity vector between each text input and prototype $k$, denoted as $\tilde{s}_k = (s_{t,k})_{t=1}^T$, which is used as the input to the DPGMM.

\subsubsection{Similarity Approximation via GMM}

Effectively capturing the most relevant text spans that align with a prototype is challenging due to the complex patterns and information in the text. We apply a Dirichlet Process Gaussian Mixture Model (DPGMM)~\citep{gorur2010dirichlet, rasmussen1999infinite} to model similarity distributions using Gaussian components, with higher similarity values highlighting important regions.

The learned Gaussian parameters correspond to high-similarity regions, thereby facilitating the identification of key relevant text parts. This enhances the model’s capability to capture complex distributions and improves interpretability by emphasizing critical text regions.

Given $\tilde{s}_k$, 
DPGMM leverages this similarity vector to determine the number of Gaussian components dynamically, focusing on the most relevant parts and effectively capturing intricate relationships. Each $\tilde{s}_k$ is approximated using up to $M$ Gaussian components:

\begin{equation}
    \tilde{s}_k = \sum_{m=1}^M \pi_m \cdot \mathcal{N}(\tilde{s}_k \mid \mu_m, \sigma_m),
\end{equation}
where $\pi_m$ denotes the mixture weight, and $\mathcal{N}(\tilde{s} \mid \mu_m, \sigma_m)$ represents the Gaussian distribution parameterized by the mean $\mu_m$ and the standard deviation $\sigma_m$.

To model each similarity distribution as a mixture of Gaussian components, we use a neural network that takes a hidden representation $h$ as input, which is derived from $\tilde{s}$ via a two-layer MLP:

\begin{equation}
    h = \text{MLP}(\tilde{s}_k),
\end{equation}

This hidden representation $h$ is then used to generate the parameters of the Gaussian mixture, including the mixture weights $\pi$, means $\mu$, and standard deviations $\sigma$, allowing the model to approximate the similarity distribution effectively.

\noindent \textbf{Means ($\mu$)} and \textbf{Standard Deviations ($\sigma$).} The parameters of the Gaussian components are computed as follows:

\begin{equation}
    \mu = \text{sigmoid}(\mathbf{W}_{\mu} h + \mathbf{b}_{\mu}) \times T,
\end{equation}
\begin{equation}
    \sigma = \exp(\mathbf{W}_{\sigma} h + \mathbf{b}_{\sigma}),
\end{equation}
where $\mu$ and $\sigma$ are the mean and standard deviation for each of the $M$ Gaussian components.

\noindent \textbf{Mixture Weights ($\pi$).} To dynamically determine the mixture weights, we employ the Stick-Breaking Process~\citep{ren2011logistic}, with the Dirichlet Process (DP)~\citep{teh2010dirichlet} implicitly implemented through the stick-breaking formulation. The DP provides a nonparametric Bayesian approach that allows the model to determine the appropriate number of components adaptively, which is crucial for handling data with unknown complexity.

We define a maximum number of Gaussian components, $M$, which represents the potential number of components for approximating the similarity distribution. The mixture weights $\pi_m$ for each component $m$ are generated as follows:

\begin{equation}
    \nu_m = \text{sigmoid}(\mathbf{W}_{\pi} h + \mathbf{b}_{\pi}),
\end{equation}
\begin{equation}
    \pi_m = \nu_m \prod_{\ell=1}^{m-1} (1 - \nu_\ell), \quad m = 1, \dots, M,
\end{equation}

Here, $\nu_m$ is computed by applying a sigmoid function to a linear transformation of the hidden representation $h$. The Stick-Breaking Process ensures that the mixture weights $\pi_m$ sum to one and adaptively determine the number of active components, enabling the model to capture complex and potentially multi-modal distributions.

\begin{table*}[t]
\centering
\caption{Performance of ProtoLens in comparison with baselines.}
\label{tab:performance_comparison}
\begin{tabular}{l@{\hskip 15pt}|@{\hskip 15pt}c@{\hskip 15pt}|@{\hskip 15pt}c@{\hskip 15pt}|@{\hskip 15pt}c@{\hskip 15pt}|@{\hskip 15pt}c@{\hskip 15pt}|@{\hskip 15pt}c}
\hline
\textbf{Model} & \textbf{IMDB} & \textbf{Amazon}  & \textbf{Yelp}  &\textbf{Hotel}  & \textbf{Steam}  \\
\hline
MPNet        & 0.846    & 0.899    & 0.950     & 0.961 &  0.913   \\
Bag-of-words & 0.877 & 0.830 & 0.908  & 0.905 & 0.844 \\
ProSeNet     & 0.863 & 0.875 & 0.932  & 0.930 & 0.834 \\
ProtoryNet   & 0.871 & 0.890 & 0.941  & 0.949 & 0.876 \\
\hline
\textbf{ProtoLens} & \textbf{0.903*} & \textbf{0.937*} & \textbf{0.962*} & \textbf{0.963*} & \textbf{0.931*} \\
\hline
\end{tabular}
\end{table*}

\begin{table}[t]
\centering
\caption{Performance of ProtoLens with different ablation settings on various datasets.}
\label{tab:ablation}

\begin{tabular}{l|c|c|c}
\hline
Dataset      & ProtoLens    & \textit{w/o Diversity} & \textit{w/o Alignment} \\

\hline

IMDB         & 0.903           & 0.882              & 0.886             \\
Amazon       &  0.937          & 0.926              & 0.927             \\
Yelp         &  0.962          & 0.931              & 0.943             \\
Hotel        &  0.963          & 0.947              & 0.953             \\
Steam        & 0.931           & 0.917              & 0.923             \\
\hline
\end{tabular}
\end{table}

\subsubsection{Span Extraction}

The learned $\mu$ can be viewed as an anchor point, representing the location with the highest probability density between the text instance and its corresponding prototype, identifying the most similar unit of the text for that prototype. We next extract the span based on the anchor, thereby capturing prototype-specific information. To achieve this, we introduce a span function, as illustrated in Figure~\ref{fig:span_function}. In this context, $\mu$ represents the anchor point, while $\sigma$ defines the range around the anchor, indicating how far from the anchor the model should be captured. The variable $x$ refers to the relative distance to $\mu$.  This function allows the model to focus on a specific area around $\mu$, with the spread of this focus determined by $\sigma$.

The span function is computed as:
\begin{equation}
    m_{\mu, \sigma}(x) =  \min(\max(\frac{1}{R} (R + \sigma - |\mu - x|), 0), 1)
\end{equation}
where $\mu$ is the anchor (the position with the highest similarity), $\sigma$ controls how far from $\mu$ the model should be captured, and $R$ is a hyperparameter that adjusts the smoothness of the span.

\subsection{Prototype Alignment}


A core requirement for delivering interpretable classifications that are understandable for humans is to make sure learned prototypes are semantically meaningful. The learned prototypes are represented as numerical embeddings, which are not inherently interpretable by human users and require further interpretation.

Hence, guidance is needed to ensure these prototype embeddings are semantically meaningful. To this end, we propose a prototype alignment mechanism that maps the prototypes to representative sentences from the training data throughout the learning process.

\paragraph{Representative Candidates.}
We begin by encoding all sentences in the training samples into embeddings. In the embedding space, we apply the k-means clustering to group the sentences based on their semantic similarity. The top 50 sentences closest to each cluster center obtained from the k-means algorithm serve as representative examples of each cluster, making them suitable candidates for aligning prototypes. 


\paragraph{Prototype Alignment.}
In Figure~\ref{fig:alignment}, we depict the prototype alignment process in ProtoLens. At one epoch during training, for each prototype with its current learned embedding $\mathbf{p}_k$, the top $3$ most similar candidate sentences (green circles) from the representative candidates are selected. These candidates are averaged to form a representative embedding $\mathbf{c}_k$ (purple cross), which encapsulates the meaning from actual training data. The prototype is then updated towards $ \mathbf{c}_k$ (orange arrow), resulting in an updated prototype $\mathbf{p}_k'$  (yellow star).





Specifically, $\mathbf{p}_k$ is updated towards $\mathbf{c}_k$ controlled by a weight factor $\omega_k$:
\begin{equation}
    \omega_k = \text{sigmoid}(\gamma \cdot (d_k - \tau)),
\end{equation}
where $d_k$ represents the Euclidean distance between $\mathbf{p}_k$ and $\mathbf{c}_k$, $\tau$ is the movement threshold and $\gamma$ controls the smoothness of the transition. 





The updated prototype $\mathbf{p}_k'$ is derived as a weighted combination of $\mathbf{p}_k$ and the movement towards $\mathbf{c}_k$:
\begin{equation}
    \mathbf{p}_k' = \omega_k \cdot (\mathbf{p}_k + \tau \cdot \mathbf{u}_k) + (1 - \omega_k) \cdot \mathbf{c}_k,
\end{equation}
where $\mathbf{u}_k$ is the unit vector pointing from $\mathbf{p}_k$ to $\mathbf{c}_k$, defined as:
\begin{equation}
    \mathbf{u}_k = \frac{\mathbf{c}_k - \mathbf{p}_k}{d_k + \epsilon},
\end{equation}
with $\epsilon$ being a small value to prevent division by zero.


If $ \mathbf{p}_k $ is far from $ \mathbf{c}_k $ (i.e., $ d_k \geq \tau $), $ \mathbf{p}_k $ will move a distance of $ \tau $ toward $ \mathbf{c}_k $. Conversely, if $ d_k \leq \tau $, $ \mathbf{p}_k $ is directly aligned with $ \mathbf{c}_k $. This process ensures that the prototypes shift toward semantically meaningful regions without abrupt changes.


\subsection{Learning Objectives}
The learning objectives of the proposed model consist of three key components that contribute to both prediction accuracy and the interpretability of the learned representations.




\subsubsection{GMM Loss}
To approximate complex similarity distributions between text samples and prototypes, we employ a Negative Log-Likelihood (NLL) loss for GMM jointly trained with the model, which is given by:
\begin{equation}
    \mathcal{L}_{\text{NLL}} = -\log (\sum_{m=1}^M \pi_m \cdot \mathcal{N}(\tilde{s} \mid \mu_m, \sigma_m) + \epsilon ),
\end{equation}
where $\pi_m$, $\mu_m$, and $\sigma_m$ are the mixture weights, means, and standard deviations of the $m$-th Gaussian component, respectively, and $\epsilon$ is a small constant added for numerical stability.

The overall loss for the GMM is defined as:

\begin{equation}
    \mathcal{L}_{\text{GMM}} = \mathbf{E}[\mathcal{L}_{\text{NLL}}] + \mathcal{L}_{\text{L1}},
\end{equation}

where an $L_1$ regularization term is introduced to promote sparsity in the mixture weights: $\mathcal{L}_{\text{L1}} = \lambda \sum_{m=1}^M |\pi_m|$, where $\lambda$ controls the regularization strength. This sparsity encourages the model to focus on a few significant Gaussian components. $\lambda$ is set to $1e^{-3}$ for all experiments. 


\subsubsection{Diversity Loss}
To promote diverse prototype representations and mitigate redundancy, we introduce a \textbf{Diversity Loss} based on cosine distance: 
\begin{equation}
    \mathcal{L}_{\text{div}} = \sum_{i \neq j} D_{ij},
\end{equation}
where $D_{ij}= 1 - \cos(\mathbf{p}_i, \mathbf{p}_j)$ be the cosine distance between prototypes. Minimizing this diversity loss enhances generalization and interpretability by maintaining a diverse set of prototypes.

\subsubsection{Overall Objective}
The final objective function for the proposed model is a weighted combination of the aforementioned loss components:

\begin{equation}
    \mathcal{L} = \text{CrossEntropy}(y, \hat{y}) + \alpha \mathcal{L}_{\text{GMM}} + \beta \mathcal{L}_{\text{div}},
\end{equation}
where  $y$ represents the true labels, $\hat{y}$ denotes the prediction, $\alpha$ and $\beta$ are hyperparameters that control the balance between accuracy, Gaussian mixture modeling, and prototype diversity. $\alpha$ and $\beta$ is set to $1e^{-1}$ and $1e^{-3}$ for all experiments, respectively. 

\section{Experiments}
In this section, we conduct comprehensive experiments to evaluate the proposed model and answer the following research questions: \textbf{RQ1}: How does ProtoLens perform in terms of classification accuracy compared to state-of-the-art (SOTA) baselines? \textbf{RQ2}: What is the quality of the model interpretations? \textbf{RQ3}: What are the effects of the proposed Prototype Alignment mechanism and Diversity loss on ProtoLens? \textbf{RQ4}: What are the impacts of different hyperparameters on ProtoLens?

\begin{figure}[t]
    \centering
    \includegraphics[width=\linewidth]{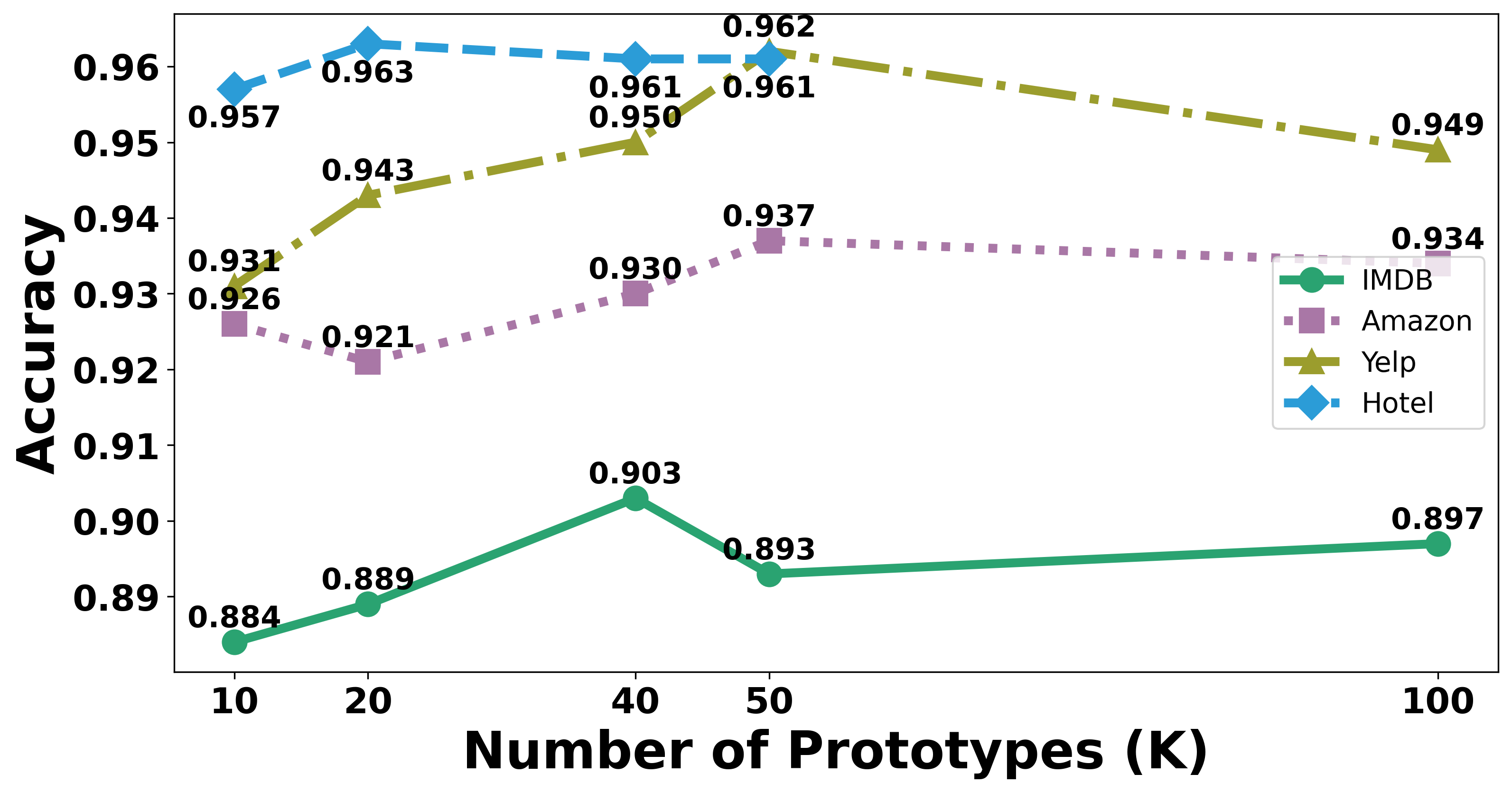}
    \caption{Performance of ProtoLens in comparison with different number of prototypes. }
    \label{fig:K}
\end{figure}

\begin{figure}[t]
    \centering
    \includegraphics[width=\linewidth]{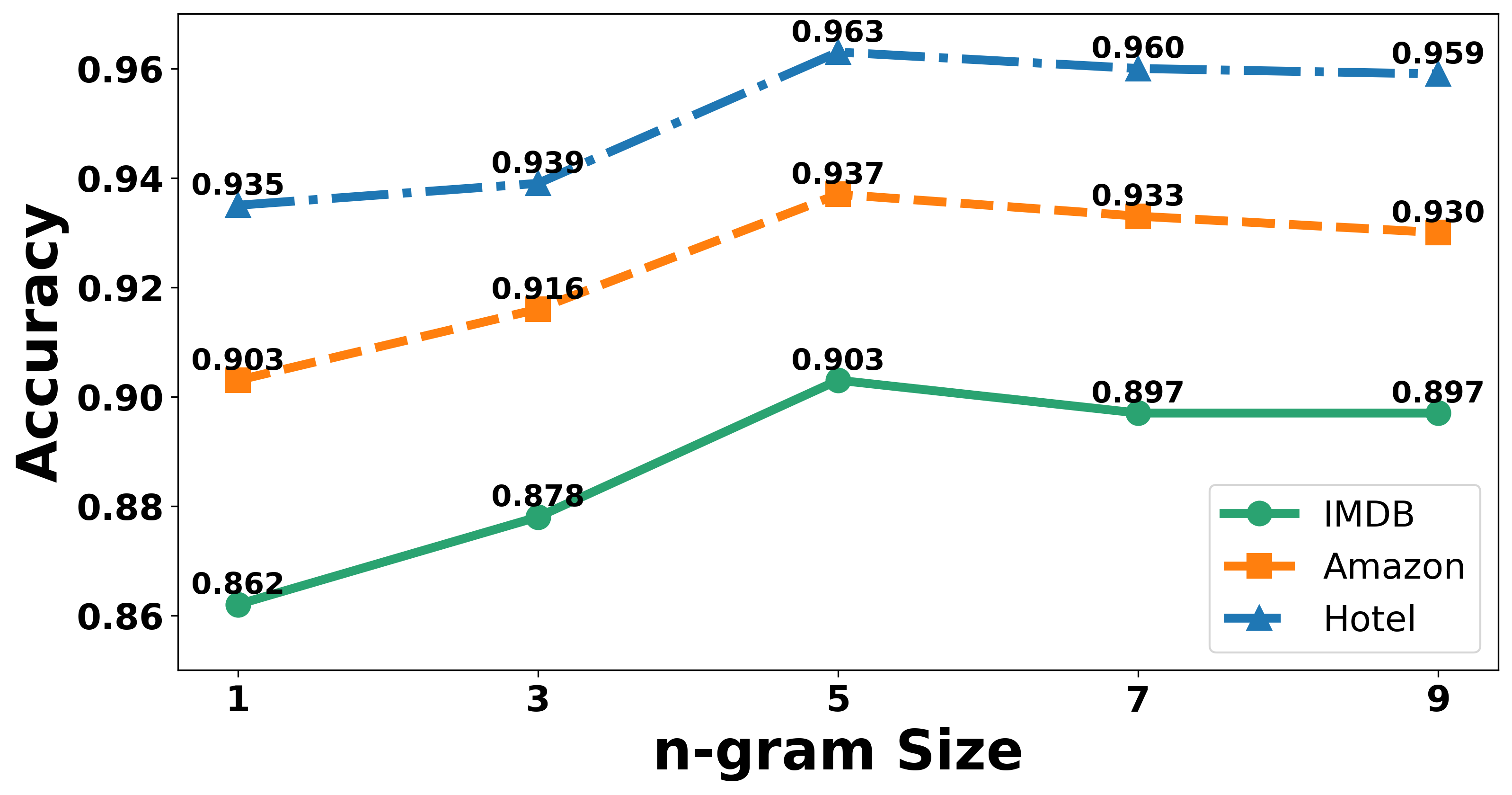}
    \caption{Performance of ProtoLens in comparison with different n-gram size. }
    \label{fig:gram}
\end{figure}

\subsection{Experimental Setup}

\noindent \textbf{Datasets.} To evaluate the performance of ProtoLens, we conducted extensive experiments using five diverse sentiment classification datasets, including IMDB, Yelp, Amazon , Hotel, and Steam. Details of these datasets are shown in Appendix~\ref{appendix:datasets}. 



\noindent \textbf{Reproducibility.} The ProtoLens model was implemented using PyTorch. For training, the prototype number $K$ is selected from $\{10, 20, 40, 50, 100\} $. The learning rate is selected from $ \{1e-4, 1e-5, 5e-5\}$, with a decay of 10\% every 10 epochs. We used the AdamW optimizer \cite{loshchilov2017decoupled} with a batch size of 16 for 25 epochs and the n-gram size is selected from $\{1, 3, 5, 7, 9\}$. The experiments were conducted on an NVIDIA A100 80GB GPU. Code and data are available at \url{https://anonymous.4open.science/r/ProtoLens-CE0B/}.

\noindent \textbf{Baselines.} We compare ProtoLens against a range of baselines, encompassing both interpretable and non-interpretable models. The interpretable baselines include ProSeNet \cite{ming2019interpretable} and ProtoryNet~\citep{hong2023protorynet}, both are prototype-based methods that provide insights into their predictions via learned prototypical representations. Additionally, we include a MPNet~\citep{song2020mpnetmaskedpermutedpretraining} and a Bag-of-Words model~\cite{zhang2010understanding} using TF-IDF representations and Logistic Regression for interpretable classification \cite{hosmer2013applied}. 



\subsection{Prediction Accuracy (RQ1)}
We evaluate the accuracy of ProtoLens against several competitive baselines, including both prototype-based and non-prototype-based methods. The results are presented in Table~\ref{tab:performance_comparison}. ProtoLens consistently achieves the highest scores, outperforming the baselines in all cases. The consistently higher performance of ProtoLens demonstrates its effectiveness and robustness across diverse domains, highlighting its superiority in leveraging fine-grained interpretability without sacrificing predictive accuracy.

\subsection{Model Interpretations (RQ2)}

The interpretability of our proposed ProtoLens is two-fold. First, ProtoLens employs prototypes that are aligned with real sentences from the training set, representing concepts with assigned weights. For a given text instance, the model reveals which concepts are present and their respective importance for the prediction, resulting in intrinsic interpretability. Second, ProtoLens extracts prototypical spans from the text that maximally activate the corresponding concept. Users can intuitively inspect these concepts by comparing the selected spans from the input with the corresponding prototypes, providing fine-grained, prototype-based interpretability.

\subsubsection{Prototype Interpretation}
 
In this section, we present an example of ProtoLens trained on the IMDB dataset with $K=10$ prototypes. Figure~\ref{fig:aligned_interpretation} showcases five randomly selected prototypes along with their aligned sentence interpretations. These prototypes span a wide range of concepts, including acting, horror elements, humor, storyline, and film execution.

What stands out is that ProtoLens achieves high accuracy while relying on concise and interpretable prototypes, often represented by short sentences. This allows for rapid and straightforward comprehension of the model’s reasoning process. Each prototype captures key characteristics of the corresponding text, providing insightful interpretations for various aspects of the movie, such as acting quality, humor, or poor execution. This feature enhances both the model's interpretability and usability, as users can easily relate the prototypes to human-understandable concepts, making the predictions more transparent. A detailed prototype interpretation for all datasets is provided in Appendix 


\subsubsection{Classification Interpretation}
When conducting classification on a text sample, ProtoLens extracts the most relevant spans from the sample for all prototypes. Similarities between spans and prototypes are then calculated to determine which concepts are activated for the sample. Last, interpretable classification is delivered based on the similarities. We present a positive example in Figure~\ref{fig:case_pos} and a negative example in Figure~\ref{fig:case_neg}, both from the IMDB dataset.

As shown in Figure~\ref{fig:case_pos}, the top three prototypes with the highest similarity scores significantly influence the classification. Prototype 0 captures the concept of a "highly entertaining flick" (similarity score 0.708, sentiment weight 0.985), Prototype 2 reflects humor with the span "crime comedy that's often very funny" (score 0.549, weight 0.247), and Prototype 5 highlights good acting with "some great actors playing these characters" (score 0.730, weight 0.931). These prototypes, focusing on entertainment, comedy, and acting, lead the model to correctly predict a "Positive" sentiment.

In contrast, Figure~\ref{fig:case_neg} shows a negative example. The text activates prototype 4, reflecting dissatisfaction with special effects, as captured in the span "problems with this film: 1 cheap special effects," with a similarity score of 0.657 and a sentiment weight of -0.956. Prototype 7 reflects frustration with the movie, highlighted by the span "ended up watching it the whole 2 hours," scoring 0.676 with a weight of -0.809. Prototype 9 captures disappointment with the lack of character development, aligned with the span "there was no character development," with a similarity score of 0.664 and weight of -0.756. These prototypes highlight negative aspects of the movie, leading the model to correctly predict the sentiment as "Negative".


\subsection{Ablation Study (RQ3)}
To demonstrate the effectiveness of the Prototype Alignment and Diversity Loss, we compare ProtoLens trained with and without these components. Prototype Alignment ensures that prototypes maintain their semantic faithfulness. The Diversity Loss encourages prototypes to be distinct, reducing redundancy in representation. The results, shown in Table~\ref{tab:ablation}, indicate that both the Prototype Alignment and Diversity Loss are essential for maintaining ProtoLens's high performance and interpretability, as their removal leads to significant declines in accuracy across datasets. A detailed analysis is provided in Appendix~\ref{appendix:ablation}.


\subsection{Hyperparameter (RQ4)}
We explored the impact of varying the number of prototypes $K$ and n-gram sizes on ProtoLens's performance, identifying dataset-specific optimal values that balance model complexity and classification accuracy. In conclusion, the optimal number of prototypes \(K\) varies by dataset, with \(K=50\) performing best for Amazon and Yelp, \(K=40\) for IMDB, and \(K=20\) for Hotel, while an n-gram size of 5 consistently yields the best results across all datasets, balancing complexity and performance. A detailed analysis is provided in Appendix~\ref{appendix:hyper}.

\section{Conclusion}
In this paper, we present ProtoLens, a prototype-based model offering fine-grained, sub-sentence level interpretability for text classification. we introduce a Prototype-aware Span Extraction module with a Prototype Alignment mechanism to ensure prototypes remain semantically meaningful and aligned with human-understandable examples. Extensive experiments across multiple benchmarks show that ProtoLens outperforms both prototype-based and non-interpretable baselines in accuracy while providing more intuitive and detailed explanations.

\section{Limitations}

While ProtoLens offers significant advancements in interpretability through prototype-based reasoning and fine-grained sub-sentence level analysis, there are several limitations to consider. First, the quality of the learned prototypes heavily depends on the training data. If the data contains inherent biases, these biases may be reflected in the prototypes, potentially leading to biased predictions or explanations. This limitation underscores the importance of careful data curation and ongoing monitoring of the model's outputs to mitigate bias.

Second, ProtoLens currently focuses on text classification tasks and has not yet been evaluated on more complex natural language processing (NLP) tasks such as machine translation or summarization. Adapting ProtoLens to these tasks may require significant architectural changes to maintain interpretability without compromising performance.

Future work could explore methods to address these limitations, such as developing techniques to automatically detect and mitigate biases, extending ProtoLens to more complex tasks, and improving the efficiency and user-friendliness of the learned interpretations.

\section{Ethics}

We have carefully considered the ethical implications of our work. ProtoLens is designed to enhance interpretability in deep neural networks, particularly for text classification tasks. By providing more transparent and intuitive explanations, ProtoLens aims to improve trust and accountability in AI systems, which is crucial in high-stakes applications such as healthcare, legal, and financial domains. 

We are committed to ensuring that the use of ProtoLens is aligned with ethical standards, promoting transparency and fairness in decision-making processes. However, as with all AI models, there is a potential risk of misuse or bias amplification if the model is trained on biased data. To mitigate this, we emphasize the importance of careful data curation and ongoing monitoring of model outputs to identify and address any unintended biases. We encourage users of ProtoLens to conduct thorough bias audits and ensure that the model is applied in a fair and responsible manner.

Furthermore, the datasets used in our experiments, including IMDB, Yelp, Amazon, Hotel, and Steam reviews, are publicly available and widely used in the research community. We have ensured that no personally identifiable information (PII) is present in the data, and that our use of these datasets complies with relevant ethical guidelines.

In conclusion, we believe that ProtoLens contributes positively to the field of interpretable AI by improving transparency and user understanding. We acknowledge the importance of continuously evaluating and mitigating potential risks to ensure that AI systems remain fair, accountable, and ethical in their applications.
\bibliographystyle{unsrtnat}
\bibliography{main}  

\clearpage
\appendix

\begin{figure}[t]
    \centering
    \includegraphics[width=\linewidth]{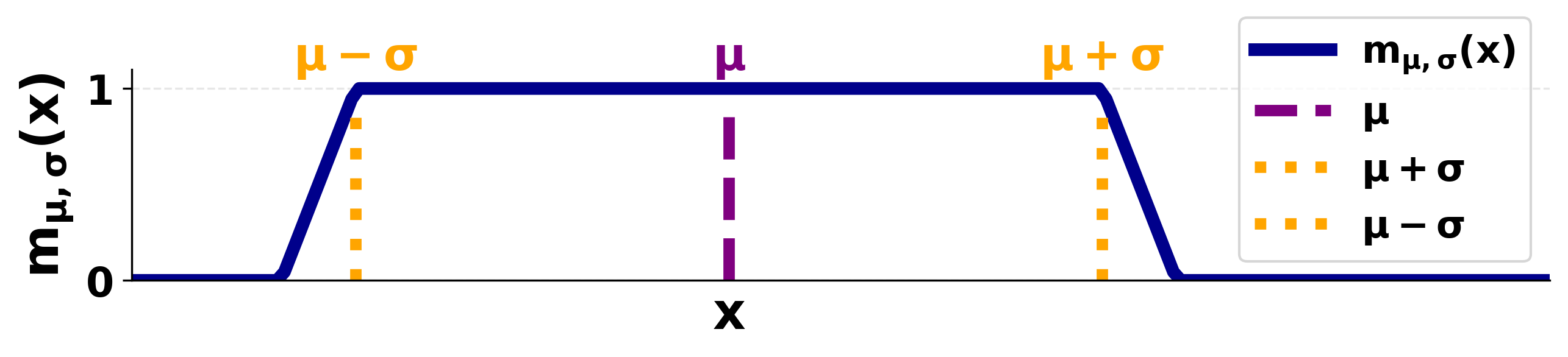}
    \caption{Span Function $m_{\mu, \sigma}(x)$. }
    \label{fig:span_function}
\end{figure}

\begin{figure}[t]
    \centering
    \includegraphics[width=\linewidth]{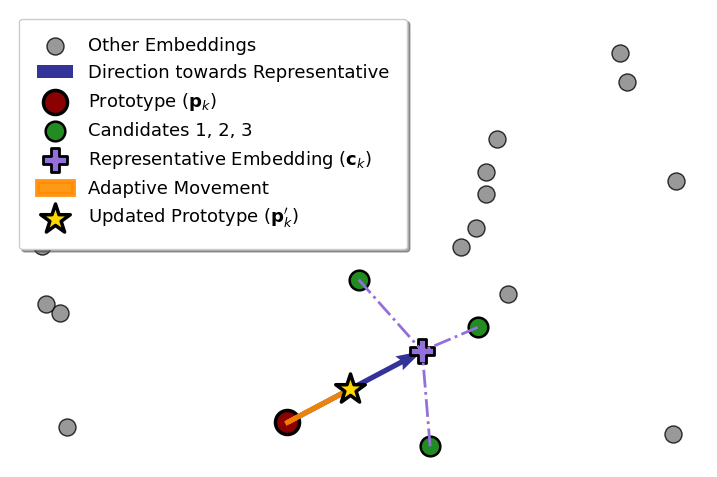}
    \caption{Prototype Alignment. }
    \label{fig:alignment}
\end{figure}

\section{Datasets}
\label{appendix:datasets}
The IMDB  dataset contains 25,000 balanced training and test samples, and follows a binary sentiment classification format. The dataset was divided into training (90\%) and validation (10\%) partitions. The Yelp Reviews dataset consists of 580,000 samples, with training and test sets of 550,000 and 30,000 samples respectively. Sentiments were binarized, treating 1-2 stars as negative and 3-4 stars as positive. The Amazon dataset was formed by selecting 30,000 random reviews, with 24,000 samples used for training and validation, and 6,000 for testing. The Hotel dataset includes 20,000 reviews evaluating 1,000 hotels, reduced to a balanced subset of 4,508 reviews (2,254 positive, 2,254 negative). Finally, the Steam Reviews dataset consists of 130,000 pre-processed reviews, balanced between positive and negative sentiments, with reviews of less than 10 characters or containing fewer than two sentences removed.

In all experiments, pre-trained embeddings from the BERT-based language model \cite{song2020mpnet} were employed to convert raw text into sentence embeddings, enabling downstream analysis.

\section{Classification Performance}
\label{appendix:classification_performance}

\section{Prototype Interpretation}
\begin{figure*}[t]
    \centering
    \includegraphics[width=\linewidth]{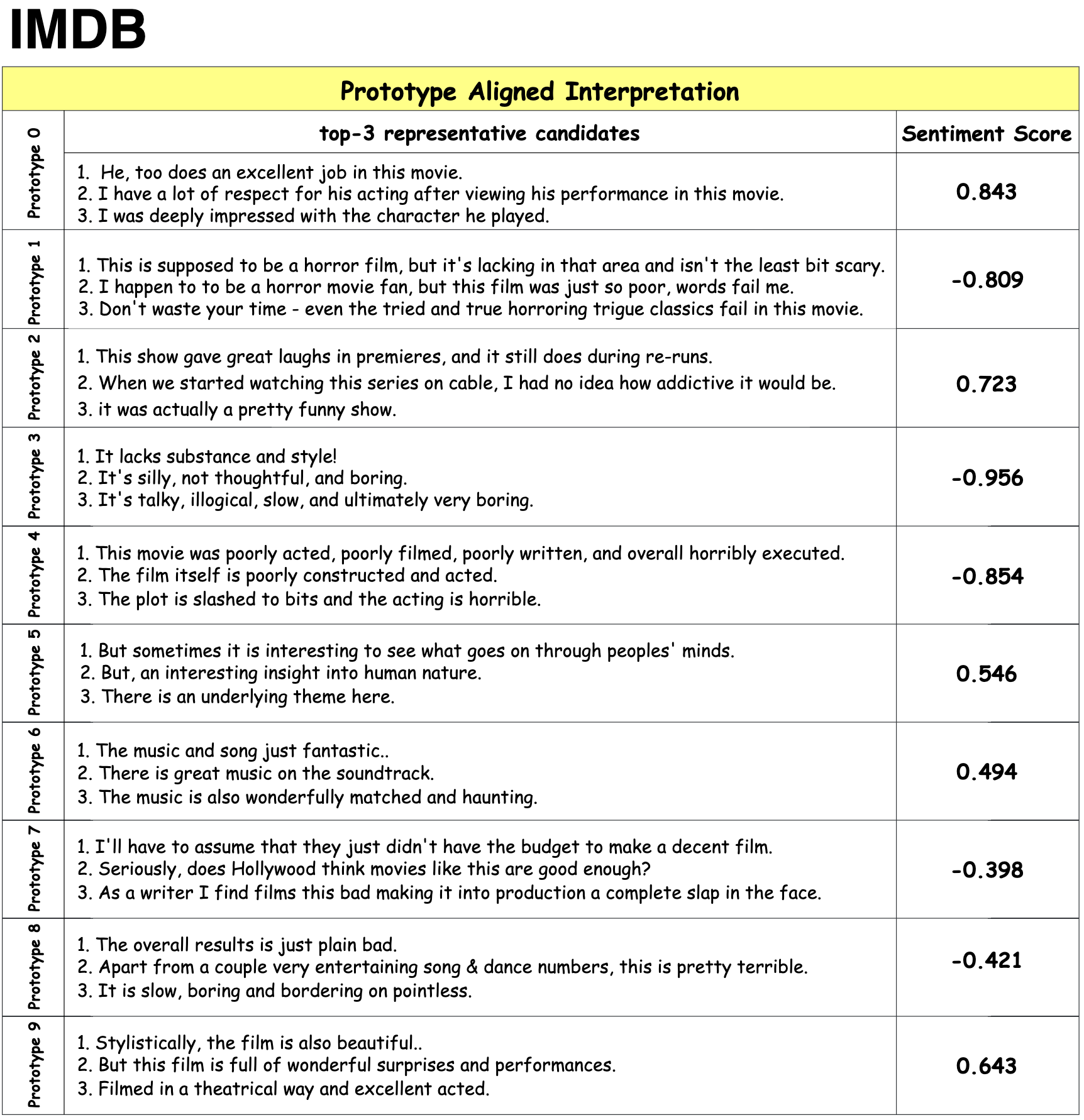}
    \caption{Aligned interpretation of prototypes with corresponding text sentences on the IMDB dataset.}
    \label{fig:aligned_interpretation_IMDB}
\end{figure*}
\begin{figure*}[t]
    \centering
    \includegraphics[width=\linewidth]{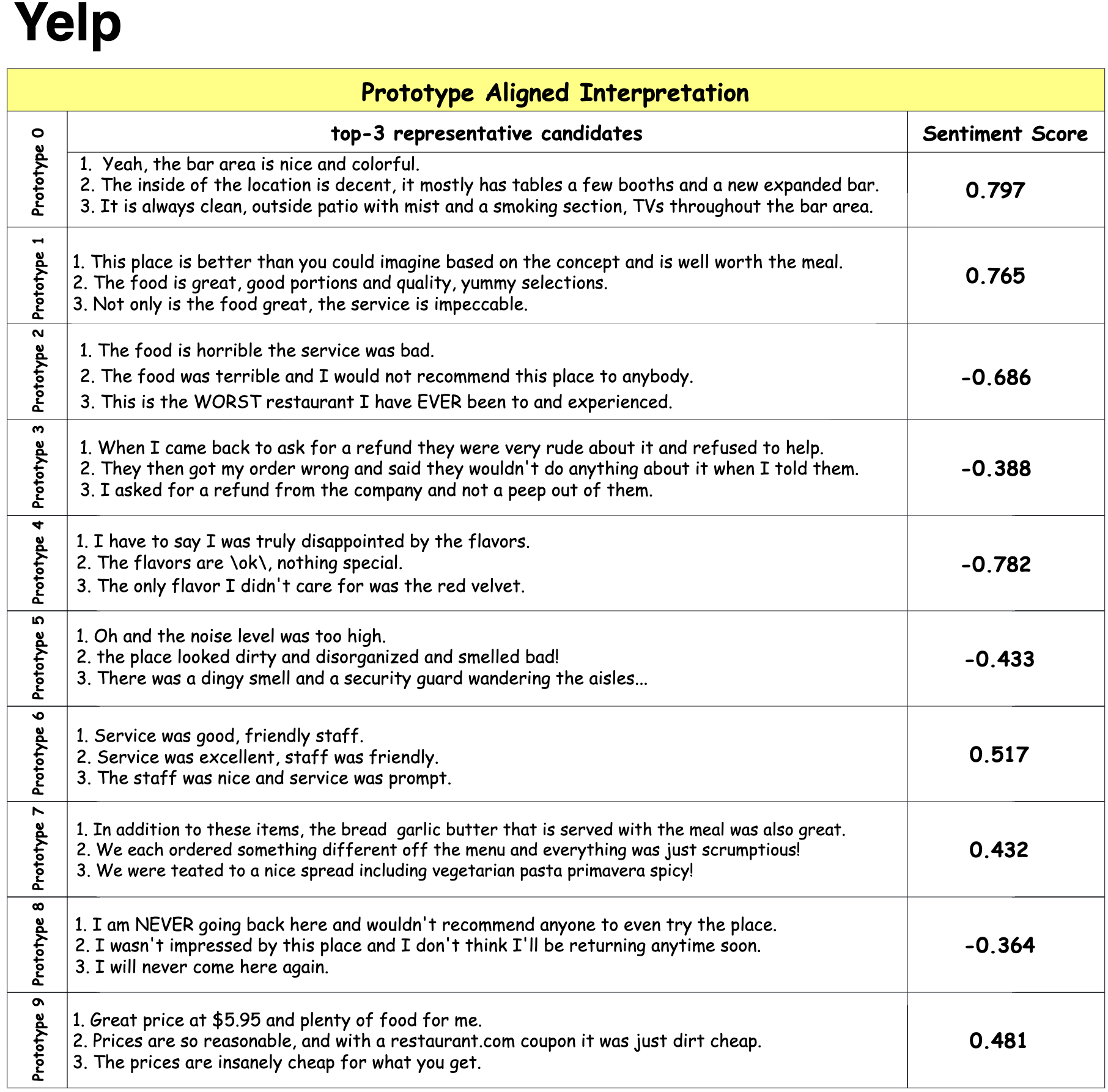}
    \caption{Aligned interpretation of prototypes with corresponding text sentences on the Yelp dataset.}
    \label{fig:aligned_interpretation_Yelp}
\end{figure*}

\begin{figure*}[t]
    \centering
    \includegraphics[width=\linewidth]{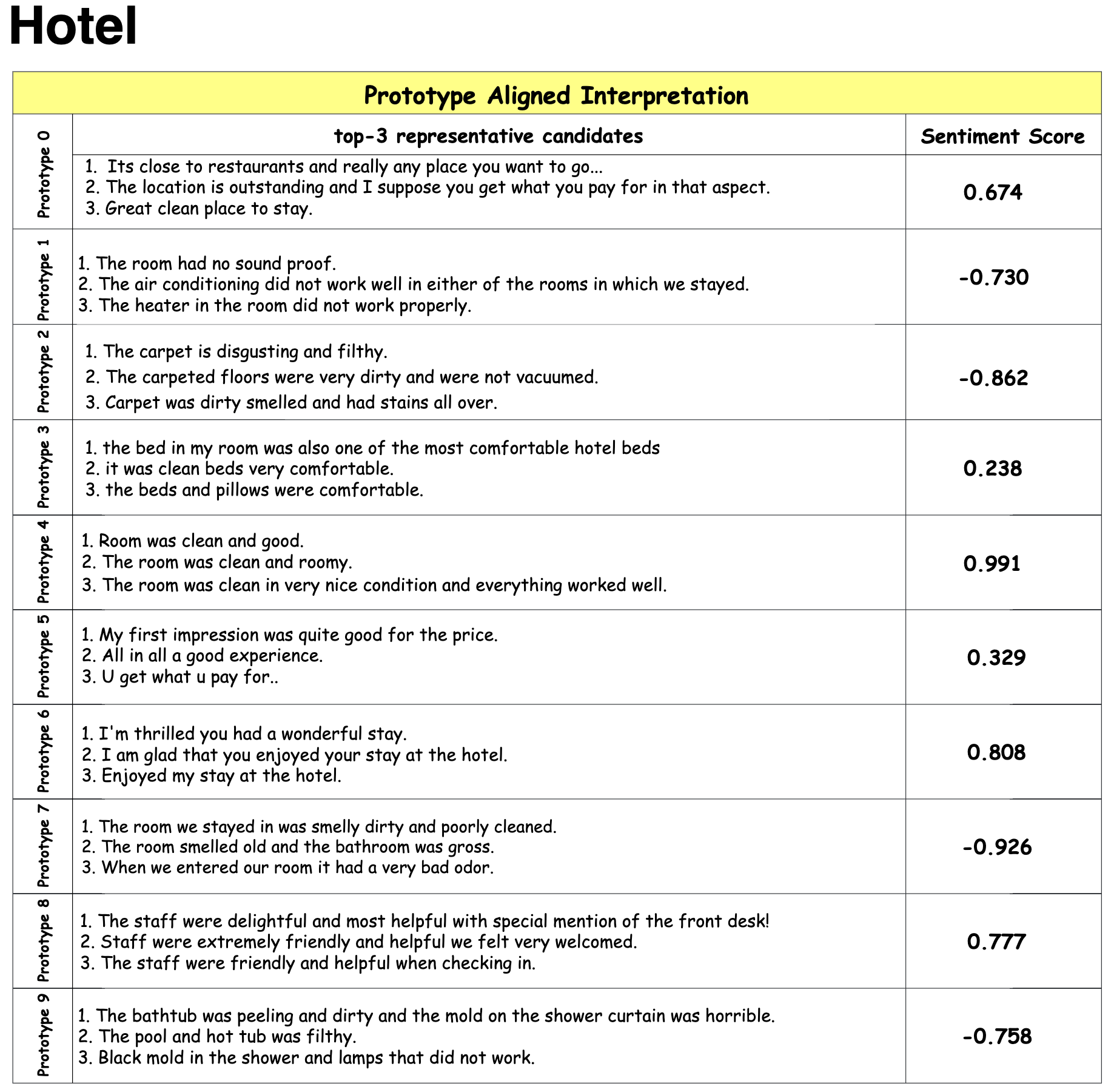}
    \caption{Aligned interpretation of prototypes with corresponding text sentences on the Hotel dataset.}
    \label{fig:aligned_interpretation_Hotel}
\end{figure*}

\begin{figure*}[t]
    \centering
    \includegraphics[width=\linewidth]{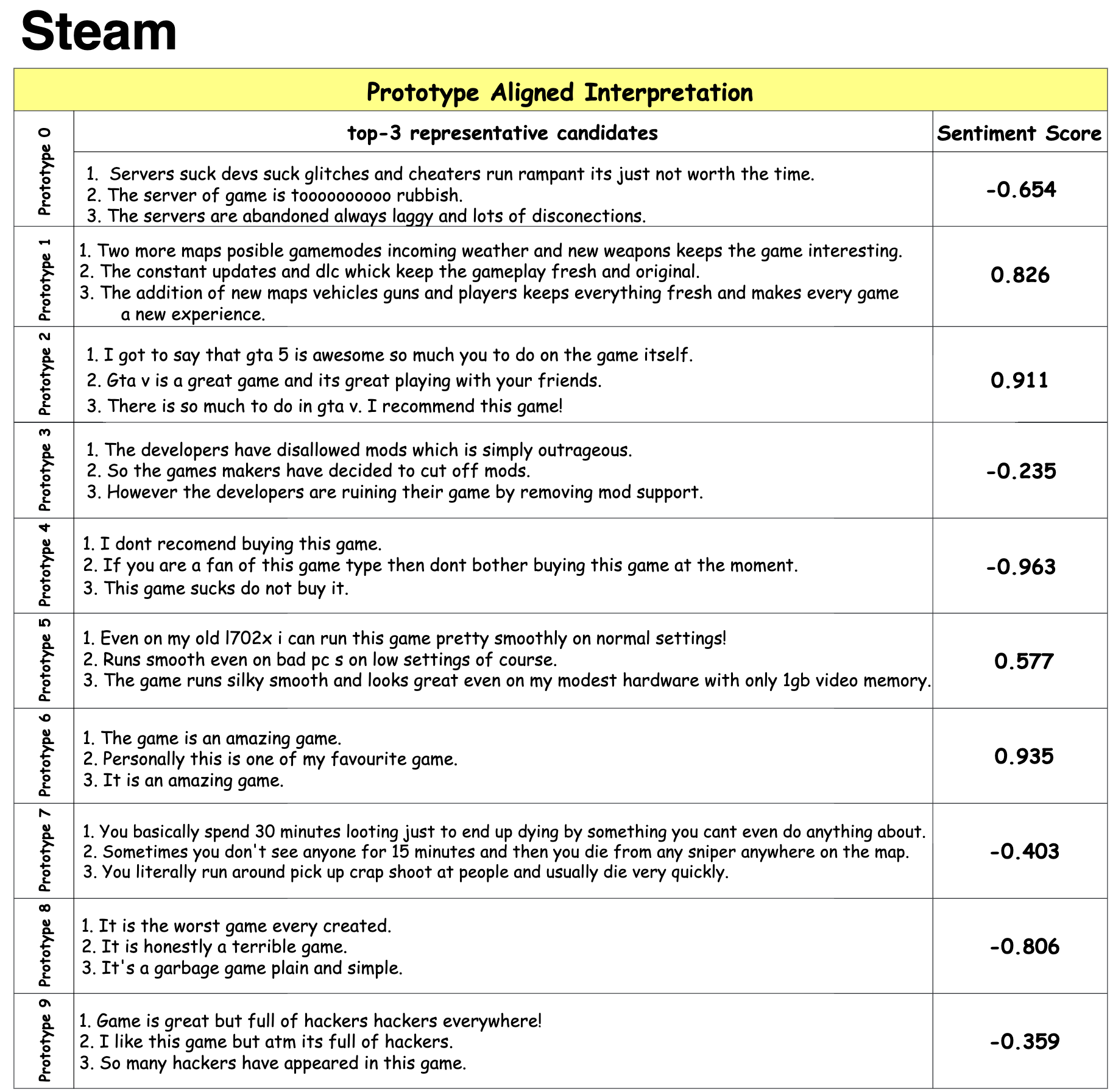}
    \caption{Aligned interpretation of prototypes with corresponding text sentences on the Steam dataset.}
    \label{fig:aligned_interpretation_Steam}
\end{figure*}

\begin{figure*}[t]
    \centering
    \includegraphics[width=\linewidth]{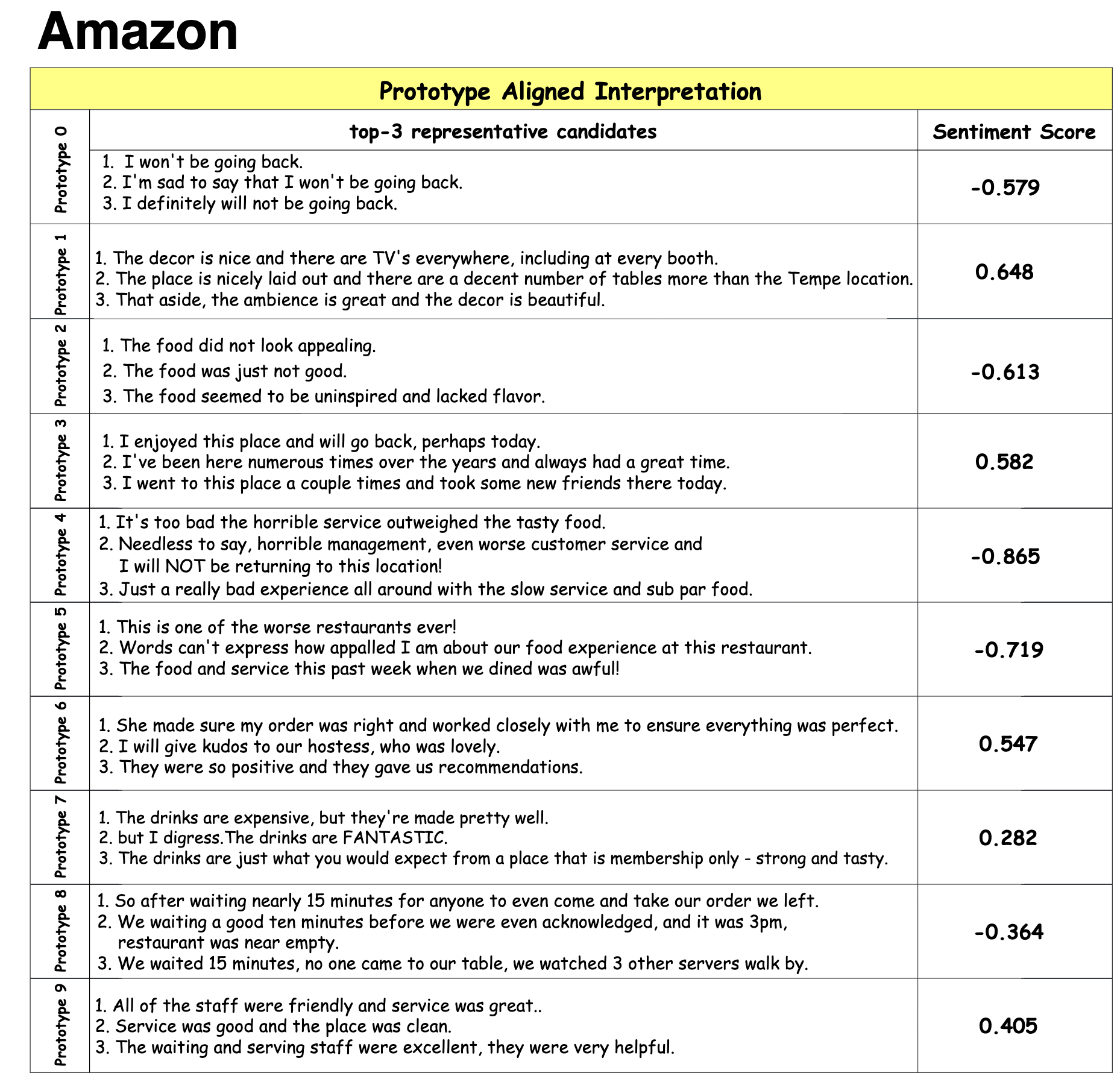}
    \caption{Aligned interpretation of prototypes with corresponding text sentences on the Amazon dataset.}
    \label{fig:aligned_interpretation_Amazon}
\end{figure*}

\section{Classification Interpretation}
\label{appendix:classification_interpretation}

\begin{figure*}[t]
    \centering
    \includegraphics[width=\linewidth]{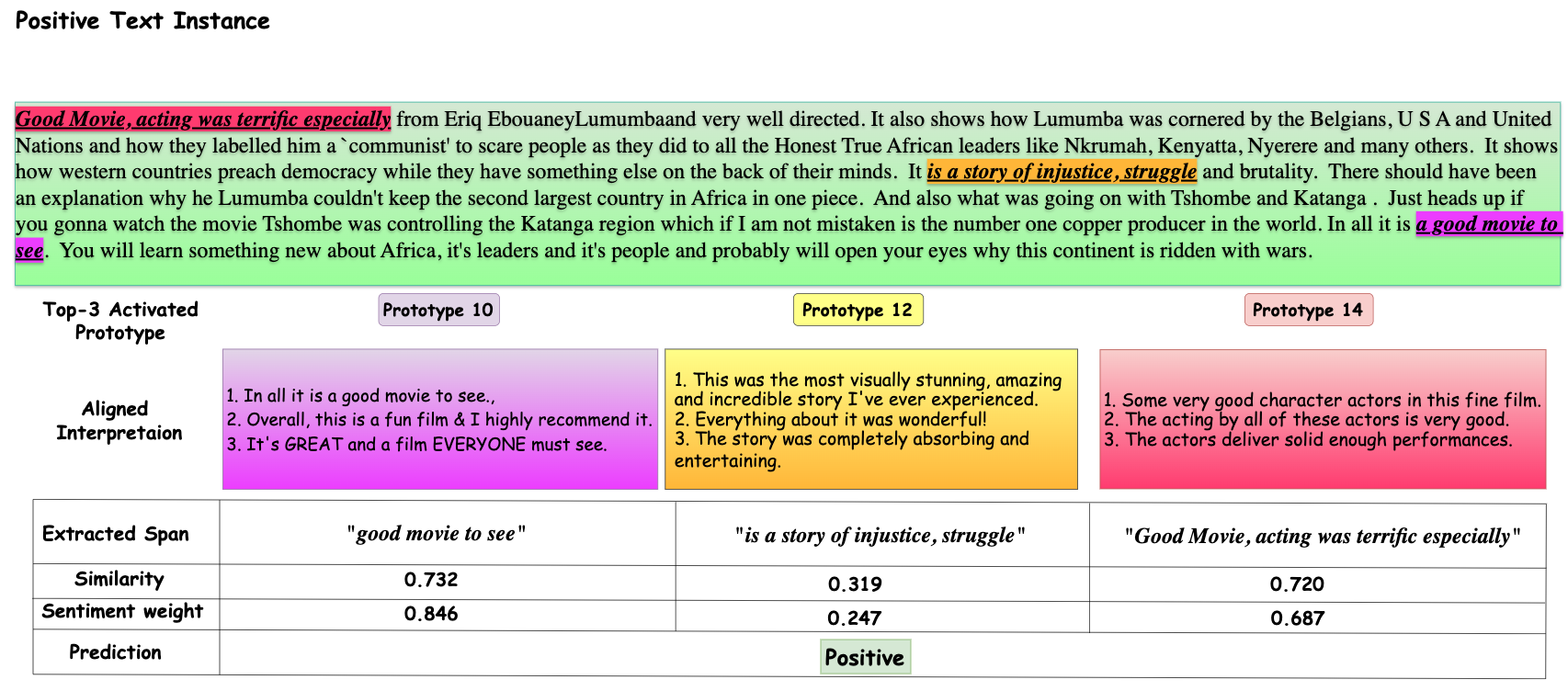}
    \caption{}
    \label{fig:pos_2}
\end{figure*}

\begin{figure*}[t]
    \centering
    \includegraphics[width=\linewidth]{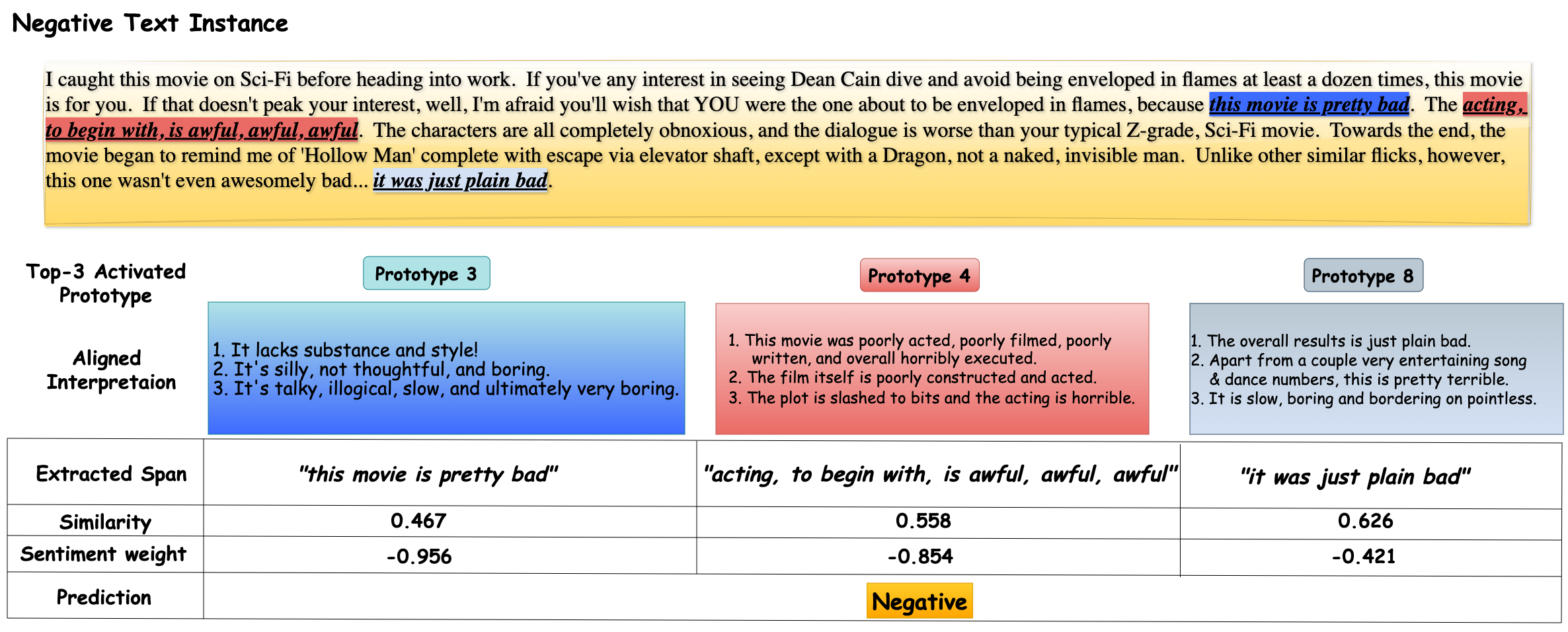}
    \caption{}
    \label{fig:neg_2}
\end{figure*}

\section{Ablation Study}
\label{appendix:ablation}
To demonstrate the effectiveness of the Prototype Alignment and Diversity Constraint, we compare ProtoLens trained with and without these components. Prototype Alignment ensures that prototypes maintain their semantic faithfulness. The Diversity Constraint encourages prototypes to capture distinct, non-redundant features, enhancing generalization and reducing redundancy in representation. The results are shown in Table~\ref{tab:ablation}.

\noindent \textbf{Impact of Diversity Constraints.} The removal of diversity constraints (\textit{w/o Diversity}) leads to a noticeable performance decline across all tested datasets, notably on IMDB (from 0.903 to 0.882), Amazon (from 0.937 to 0.926), Yelp (from 0.962 to 0.931) and Hotel (from 0.963 to 0.947). This indicates that the diversity loss plays a crucial role in encouraging distinct and varied prototype representations, which helps the model generalize better across different data points. The drop in accuracy suggests that when prototypes become more redundant, they lose their ability to represent the diversity in the dataset, limiting the model’s interpretability and performance.

\noindent \textbf{Impact of Prototype Alignment.} The ablation results for removing prototype alignment (\textit{w/o Alignment}) show a decline in performance, particularly on the Yelp dataset (from 0.963 to 0.943), highlighting the importance of prototype alignment. Aligning prototypes with representative embeddings ensures they remain semantically meaningful, leading to more accurate and interpretable predictions. The slight performance drop across other datasets, such as IMDB and Amazon, further emphasizes that the adaptive update process enabled by prototype alignment promotes more stable and reliable learning, improving the model’s interpretability and accuracy.

\section{Hyperparameter}
\label{appendix:hyper}
\noindent \textbf{Effect of K.} The number of prototypes, denoted by \(K\), plays a crucial role in determining the balance between model interpretability and classification performance. As shown in Figure~\ref{fig:K}, increasing \(K\) generally leads to improved accuracy across most datasets, with the exception of some slight fluctuations. For instance, in the IMDB dataset, increasing \(K\) from 10 to 40 boosts the performance from 0.884 to 0.903, while for the Yelp dataset, a similar increase elevates the accuracy from 0.931 to 0.950. The improvements plateau or slightly decrease for higher values of \(K\), suggesting diminishing returns beyond a certain point.

The optimal value of \(K\) appears to be dataset-dependent. For example, \(K=50\) yields the best performance on the Amazon and Yelp datasets with 0.937 and 0.962, respectively, while \(K=40\) provides the best performance on the IMDB dataset (0.903). Meanwhile, for the Hotel dataset, \(K=20\) achieves the highest accuracy at 0.963. This variation indicates that the ideal number of prototypes may depend on the complexity and size of the dataset.

Overall, increasing \(K\) allows the model to capture more fine-grained patterns by using a larger set of prototypes, but setting \(K\) too high may introduce unnecessary complexity without substantial accuracy gains. Thus, choosing \(K\) involves a trade-off between maintaining a manageable number of interpretable prototypes and achieving high predictive performance.

\noindent \textbf{Effect of n-gram.} An n-gram is a hyperparameter that determines the granularity of text division. As shown in Figure~\ref{fig:gram}, an n-gram size of 5 achieves the best performance across all datasets, with notable improvements on IMDB (0.903), Amazon (0.937), and Hotel (0.963), indicating that $n=5$ is the optimal n-gram size, providing the best trade-off between incorporating sufficient context and avoiding unnecessary complexity. For smaller n-gram sizes (e.g., $n=1, 3$), performance is slightly lower, likely due to the model's limited ability to capture broader contextual information. On the other hand, a larger n-gram size ($n=7, 9$) does not yield improved performance and even leads to a decrease in accuracy on all datasets, as seen with IMDB and Amazon. This suggests that including too large of a n-gram introduces noise, which results in slight performance degradation.

\end{document}